\pdfoutput=1

\documentclass[11pt]{article}

\usepackage[final]{coling}

\usepackage{times}
\usepackage{latexsym}

\usepackage[T1]{fontenc}

\usepackage[utf8]{inputenc}

\usepackage{microtype}

\usepackage{inconsolata}

\usepackage{graphicx}
\usepackage{tikz}
\usetikzlibrary{arrows.meta, positioning}
\usepackage{booktabs}
\usepackage[linesnumbered,ruled,vlined]{algorithm2e}
\usepackage{amsmath}
\usepackage{amssymb}
\usepackage{CJKutf8}
\usepackage{tabularx}
\usepackage{caption}
\usepackage{float}
\usepackage{array}
\usepackage{ragged2e}

\title{PANDA -- Paired Anti-hate Narratives Dataset from Asia:\\Using an LLM-as-a-Judge to Create the First Chinese Counterspeech Dataset}

\author{
  \textbf{Michael Bennie}\textsuperscript{1},
  \textbf{Demi Zhang}\textsuperscript{1},
  \textbf{Bushi Xiao}\textsuperscript{1},
  \textbf{Jing Cao}\textsuperscript{2},
  \\
  \textbf{Chryseis Xinyi Liu}\textsuperscript{1},
  \textbf{Jian Meng}\textsuperscript{1},
  \textbf{Alayo Tripp}\textsuperscript{1}
  \\
  \\
  \textsuperscript{1}University of Florida, United States  \\
  \textsuperscript{2}Lingua \\
  \small{
    \textbf{Correspondence:} \href{mailto:michaelbennie@ufl.edu}{michaelbennie@ufl.edu}, \href{mailto:zhang.yidan@ufl.edu}{zhang.yidan@ufl.edu}
  }
}

\begin{document}
\maketitle
\begin{abstract}

Despite the global prevalence of Modern Standard Chinese language, counterspeech (CS) resources for Chinese remain virtually nonexistent. To address this gap in East Asian counterspeech research we introduce a corpus of Modern Standard Mandarin counterspeech that focuses on combating hate speech in Mainland China. This paper proposes a novel approach of generating CS by using an LLM-as-a-Judge, simulated annealing, LLMs zero-shot CN generation and a round-robin algorithm. This is followed by manual verification for quality and contextual relevance. This paper details the methodology for creating effective counterspeech in Chinese and other non-Eurocentric languages, including unique cultural patterns of which groups are maligned and linguistic patterns in what kinds of discourse markers are programmatically marked as hate speech (HS).  In our analysis of the generated corpora, we provide strong evidence for the lack of open-source, properly labeled Chinese hate speech data and the limitations of using an LLM-as-Judge to score possible answers in Chinese. Moreover, the present corpus serves as the first East Asian language based CS corpus and  provides an essential resource for future research on counterspeech generation and evaluation.\footnote{The dataset can be found at \href{https://github.com/michaelbennieUFL/PANDA}{github.com/\\michaelbennieUFL/PANDA}}

\textit{\textbf{Warning:} The below text contains vulgar and oftentimes offensive speech. Any counterspeech or hate speech is used for exemplary purposes and doesn't necessarily reflect the views of any researcher involved.}

\end{abstract}

\section{Introduction}

Hate speech is typically characterized as any form of communication that demeans a specific group of people based on attributes like race, ethnicity, gender, sexual orientation, or religion \citep{de-gibert-etal-2018-hate}. While HS may constitute a small proportion of social media content, its impact is significant, affecting nearly one-third of the population \citep{vidgen-etal-2019-challenges}. The proliferation of hate speech on social media platforms has become a significant societal concern. While traditional approaches to mitigating HS have focused on content removal and moderation, these methods often raise concerns about freedom of speech. In response, counterspeech has emerged as a promising alternative strategy to combat HS while preserving free expression \citep{poudhar-etal-2024-strategy}.

Counterspeech, defined as communication that aims to counteract potential harm caused by other speech, has shown effectiveness in real-world studies \citep{https://doi.org/10.1111/phc3.12890}. However, the manual creation of CS is time-consuming and challenging to scale given the volume of HS online. This has led to increased interest in automated CS generation using NLP techniques.

\vspace{1em}
\noindent \textbf{Our contributions}
\begin{itemize}

\item We generate the first Chinese counterspeech dataset specifically designed for combating hate speech online. This resource fills a crucial gap in the field, as most existing datasets focus on English or other Western languages.


\item We introduce and evaluate novel metrics for assessing the quality and effectiveness of generated Chinese counterspeech, addressing the limitations of existing evaluation methods in this domain.

\item We implement a comprehensive annotation scheme based on established CS strategies, adapting them for the Chinese cultural and linguistic context.

\end{itemize}

\section{Background}


\subsection{Hate Speech and Counterspeech}


Counterspeech has gained traction as an alternative to content removal. Studies have demonstrated the efficacy of CS in enhancing online discourse quality and reducing the prevalence and impact of hateful behavior \citep{doi:10.1177/20563051211063843}. However, it's important to note that the effectiveness of CS can vary significantly depending on the context and specific strategies employed. For example, the quantity of training data available to train an LLM on a specific language will predict the robustness of its generative function. 

\subsection {Datasets and Annotation}

Several datasets have been developed to support research in CS generation. \citet{fanton-etal-2021-human} presented a dataset of 5,003 English HS/CS pairs covering multiple targets of hate, created using a combination of language model generation and expert review. \citet{chung-etal-2019-conan} annotated the CONAN dataset with response types using non-expert annotators.

Although there are multiple HS/CS datasets in English, both Chinese HS and CS resources are insufficient. Among six publicly available Chinese HS datasets without CS (see Table \ref{tab:dataset-summary}), merely four are readily accessible for research purposes, with varying annotation schemes and focuses. Furthermore, Chinese datasets often suffer from quality inconsistencies due to several unique challenges in the Chinese context: the prevalence of coded language and internet slang that obscures hateful content, complex linguistic variations across different Chinese-speaking regions, and social media censorship that affect data collection. These factors make it particularly challenging to obtain high-quality datasets, as annotators must possess not only linguistic expertise but also deep cultural knowledge to accurately identify and categorize HS.

%

\subsection {Counterspeech Strategies}

Several studies have identified and categorized effective CS strategies. \citet{chung2023understandingcounterspeechonlineharm} conducted a systematic review, identifying eight strategies used in social sciences and real-world policy-driven campaigns. These strategies include presenting facts to counter misinformation and using humor or satire to diffuse hostility. Expressing empathy or support for the targets of HS is another approach, as is highlighting hypocrisy or inconsistencies in hateful arguments. Additionally, questioning the logic or assumptions underlying HS, denouncing hateful speech without attacking the speaker, and offering alternative perspectives or narratives are also effective. Finally, appealing to shared values or common ground is often used to foster understanding. 
The effectiveness of these strategies can be highly context-dependent, emphasizing the need for nuanced approaches to CS generation and evaluation.

\subsection{Automated Counterspeech Generation}

Counterspeech offers several advantages over traditional content moderation approaches. First, it upholds the principles of free expression by engaging with problematic content rather than censoring it \citep{DBLP:journals/corr/abs-2106-01625}. Second, CS is not bounded by the often arbitrary definitions of hate speech used by different platforms and can be more easily adapted to be used across different platforms. Third, it creates opportunities for education and constructive dialogue, potentially addressing the root causes of hate speech.

Recent advances in NLP, particularly in large language models, have opened new possibilities for automated CS generation. Early work by \citet{qian-etal-2019-benchmark} explored various approaches, including sequence-to-sequence models, variational autoencoders, and reinforcement learning for counterspeech. More recent studies have focused on how large pretrained language models perform in both fine-tuned and zero-shot settings for counterspeech. \citet{tekiroglu-etal-2022-using} present a comprehensive comparative study on using several pre-trained Transformer-based LMs (e.g., GPT-2, DialoGPT, and BART) for generating English counter narratives. They find that autoregressive models combined with certain decoding schemes often outperform others in producing specific, non-generic responses. 

Similarly, \citet{saha-etal-2024-zero-shot} investigate zero-shot counterspeech generation using popular LLMs such as GPT-2, DialoGPT, ChatGPT, and FlanT5. They show that ChatGPT consistently generates strong counterspeech responses even in zero-shot scenarios, although certain models have higher toxicity with larger parameter sizes. Their findings underscore the importance of prompt engineering and model selection when developing robust counterspeech systems.

Earlier fine-tuning approaches by \citet{pranesh2020towards} and \citet{tekiroglu-etal-2022-using} demonstrated promising results for counterspeech, but they often struggled with producing diverse, high-quality responses. More recent work on zero-shot and few-shot settings \citep{saha-etal-2024-zero-shot} attempts to mitigate these limitations via better prompting strategies, model ensembles, or post-processing. Nonetheless, generating counter-narratives that are contextually grounded, non-repetitive, and culturally sensitive remains challenging. As such, additional innovation is required to  enhance diversity, relevancy, and alignment with community guidelines.

\subsection{Current Evaluation Metrics}

Evaluating the quality and effectiveness of generated counterspeech with automatic evaluation tools remains a significant challenge. The current study uses a combination of LLM and traditional NLP metrics:

\begin{table*}[ht]
    \centering
    \begin{tabular}{lcccc}
        \toprule
        \textbf{Datasets} & \textbf{Open Source\footnotemark{}} & \textbf{Total Instances} & \textbf{HS/Offensive Speech} & \textbf{Non-HS} \\
        \midrule
        \textbf{COLD} \cite{cold} & Yes & 37,480 & 18,041 & 19,439 \\
        \textbf{SWSR} \cite{SWSR} & Yes  & 8,969  & 894 & 8,075 \\
        \textbf{CHSD } \cite{CHSD} & Yes  & 17,430  & 7,485 & 9,945 \\
        \textbf{CDIAL} \cite{CDIAL-BIASDATASET} & No  & 28,343  & 7,233 & 21,110 \\
        \textbf{ToxiCN} \cite{toxiCN} & No  & 12,011  & 6,461 & 5,550 \\
        \textbf{Political} \citep{TaiwanHateSpeech} & No  & 315,795  & 16,976 & 298,819 \\
        \midrule
        \textbf{Used In Preprocessing} &Yes&26,420&26,420&0\\
        \bottomrule
    \end{tabular}
    \caption{Statistics of available corpora, showing the total number of instances of data, the number of instances of data that could be labeled as possible hate speech, and the number of instances of data of non-hate speech. For the current study, it only included instances of potential hate-speech from open-source corpora.}
    \label{tab:dataset-summary}
\end{table*}

\begin{itemize}
   
 \item JudgeLM: A LLM-based ranking method for evaluating automatic counter-narrative generation \citep{zubiaga2024llmbasedrankingmethodevaluation}.
 \item BLEU: Measures token overlap between predictions and references \citep{papineni2002bleu}.
 \item ROUGE-L: Computes sentence-level structure similarity and longest co-occurring n-grams \citep{lin2004rouge}.
 \item BERTScore: Calculates token-level similarity using contextual embeddings \citep{DBLP:journals/corr/abs-1904-09675}.
 \item Novelty: Measures the proportion of non-singleton n-grams in generated text that do not appear in the training data \citep{Wang2018SentiGANGS}.
 \item Genlen: The average length of generated predictions.\end{itemize}
These metrics aim to provide a more comprehensive evaluation of CS quality, addressing aspects such as relevance, diversity, and effectiveness in countering hate speech.

\section{Methodology}

\begin{figure}[ht]
    \centering
    \resizebox{0.5\textwidth}{!}{
        \begin{tikzpicture}[
            node distance=0.3cm and 0.2cm,
            every node/.style={rectangle, rounded corners, draw=black, align=center, minimum width=1cm, minimum height=1cm},
            arrow/.style={-{Stealth}, shorten >=1pt, shorten <=1pt}
            ]

        \node (a) at (0,0) {\textbf{Original HS Data}\\(SWSR, COLD, CHSD)\\63,879 Examples};
        \node (b) [below=of a] {\textbf{Pre-Processing}\\(Keeping Hate-Labeled Entries)\\26,420 Examples of HS};
        \node (c) [below=of b] {\textbf{LLM-Assisted Rating}\\(Labeling Hate Level)\\26,420 Examples of HS};
        
        \node (d) [below=of c] {\textbf{Selecting HS}\\2,974 Examples of HS};
        \node (e) [below=of d] {\textbf{AI Generate Answers}\\17,844 Examples of CS};
        
        \node (f) [right=of a, xshift=3cm] {\textbf{Round Robin Selection}\\11,896 Examples of CS};
        
        \draw [arrow] (a) -- (b);
        \draw [arrow] (b) -- (c);
        \draw [arrow] (c) -- (d);
        \draw [arrow] (d) -- (e);
        
        \draw [arrow] (e.east) to[out=0,in=-180] (f.west);
        
        \node (g2) [below=of f] {\textbf{A$_2$}};
        \node (g1) [below=of f, left=of g2] {\textbf{A$_1$}};
        \node (dots) [below=of f, right=of g2] {\textbf{...}};
        \node (g_last) [below=of f, right=of dots] {\textbf{A$_n$}};
        
        \draw [arrow] (f) -- (g1);
        \draw [arrow] (f) -- (g2);
        \draw [arrow] (f) -- (dots);
        \draw [arrow] (f) -- (g_last);
        
        \node (h) [below=of g2] {\textbf{Combined Output}\\2,974 Annotated Pairs of HS and CS };
        
        \draw [arrow] (g1) -- (h);
        \draw [arrow] (g2) -- (h);
        \draw [arrow] (dots) -- (h);
        \draw [arrow] (g_last) -- (h);
        
        \end{tikzpicture}
    }
    \caption{Proposed Data Processing Pipeline for Creating the Chinese Counterspeech Corpus. A$_1$ through A$_n$ refer to $n$ annotators that participated in this project.}
    \label{fig:data_processing}
\end{figure}
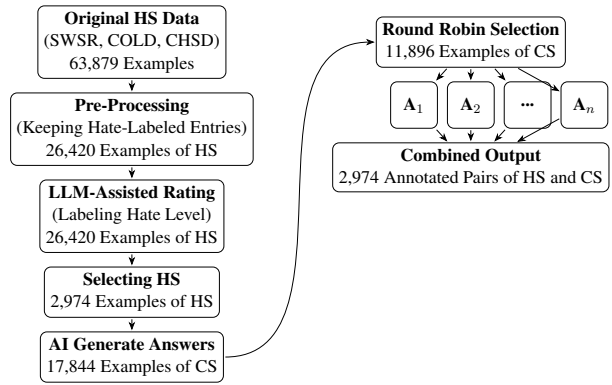

This section provides an overview of the targets we set when making this dataset (\ref{METHODS:Goals}), the sourcing of data (\ref{METHODS:Sources}), the pre-processing of data (\ref{METHODS:Preprocess}), generation of CS (\ref{METHODS:CSGen}), and  annotation methods (\ref{METHODS:Annotations}). Finally, we also provide statistics relating to the dataset and rating (\ref{METHODS:Stats}). A graphical overview is provided in Figure \ref{fig:data_processing}.

\subsection{Goals/Requirements}\label{METHODS:Goals}

We aim to achieve the following objectives:

\begin{itemize}
    \item \textbf{Creation of the First East Asian HS-CN Dataset.} 
    During our review of existing datasets, we identified significant gaps in Chinese counterspeech (CS) resources. Although datasets like \citet{cold} and \citet{CDIAL-BIASDATASET} include instances labeled as 'anti-bias', their scope and definitions do not align with the specific focus of CS research. These datasets adopt a broader concept of 'anti-bias', encompassing content that promotes fairness and addresses various forms of offensive language rather than specifically targeting hate speech. Our work addresses this gap by creating a dataset that exclusively targets hate speech and counter speech, providing a more focused resource for CS research.

    \item \textbf{Paired Structure.} 
    A notable limitation of previous datasets is the absence of a paired structure that directly links CS responses to specific instances of hate speech. In contrast, English-language datasets such as \cite{CONAN} have demonstrated the value of this framework in facilitating precise and contextual analyses of intervention strategies. Our dataset introduces this paired structure for the first time in the Chinese context, explicitly mapping CS responses to their corresponding hate speech instances.

    \item \textbf{Freely Usable.} 
    All hate speech data collected for our dataset originate from open-source repositories. Additionally, we have released our model and the generated data under a permissive GPL license. This ensures that the generated and annotated data can be freely utilized in both commercial and non-commercial projects, promoting wider accessibility and application in various research and practical initiatives.
\end{itemize}

\subsection{HS Sources}\label{METHODS:Sources}

To the best of the authors' knowledge, there have only been six published HS datasets in the literature. This data was summarized in Table \ref{tab:dataset-summary}. 

Three corpora (\citep{toxiCN},\citep{CDIAL-BIASDATASET} and \citep{TaiwanHateSpeech}) were later removed from the dataset due to restrictive licensing from them. What was left were 3 open-source datasets.

The COLDataset contains over 30,000 instances that are labeled either safe or offensive and, further, contains fine grained labels for each category \cite{cold}. The dataset was chosen due the fact that, under a cursory look, many, but not all, of the statements labeled offensive were in-fact hate speech. The second dataset used was `SexComment.csv' from SWSR. This file focuses on finding and labeling sexist comments and also contains subcategories for the type of comment and whether it is targeted at an individual or a group \cite{SWSR}.  We decided to include this dataset to increase representation of sexist hate-speech in the database. The last dataset included was from CHSD which is actually a preprocessed dataset of HS that comes from COLD, CDIAL, and SWSR \cite{CHSD}. 

\footnotetext{This paper used the Open Source Initiative's definition of open source which can be found at  \href{https://opensource.org/osd}{opensource.org/osd}.}

\begin{figure}
            \centering
            \includegraphics[width=1\linewidth]{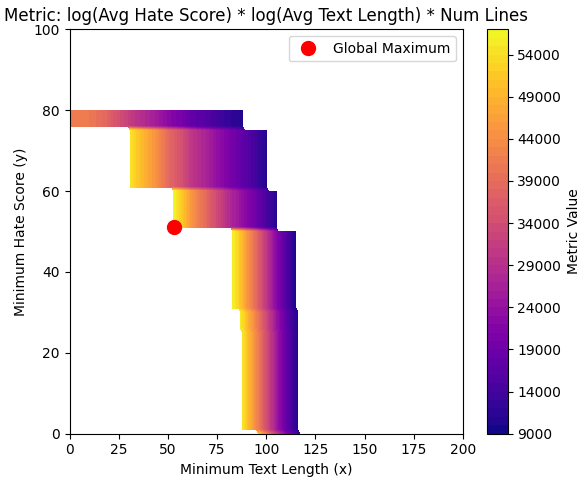}
            \caption{The scoring heat-map based on different combinations of minimum hate-speech score (y) and minimum length of each string (x).}
            \label{fig:optimizedLineCount}
\end{figure}

\subsection{Filtering of Data}\label{METHODS:Preprocess}

The initial three corpora included entries that were labeled non-hate speech. In order to avoid the unnecessary computational cost of attempting to generate CS for non-HS sentences, we initially used some commands to filter out any rows that aren't considered HS by the corpora. For CSHD, we removed any rows where `label' equals `0.' Likewise, for the COLD dataset, we kept only the data that had a label of `1' and a sub label of `1' (attacking individuals)  or `2' (attacking groups). For SWSR, we keep only the instances that had a label of `1' (sexist) and all sub-categories except for `MA' (micro aggressions) as we believed that it was harder to determine which answers counted as a hate-speech.

Once completing the first round of pre-processing, we hand annotated 19 instances of hate-speech and scored them from 0 to 100. Then, we employed a model-in-the-loop collection scheme similar to what was described in \citet{model-in-loop}. The model that we used to discriminate between non-HS and HS was based off of Llama-3.1 Instruct with 70 billion parameters. 

We then use the scores given by the LLM and the text length to optimize over the set of possible subsets of hate-speech. As we wanted to have a subset that balances between high average hate score and a high average text length, we choose the metric of $log(Average  Hate Score)* log(Average Text Length)*Num Instances$. We limited the range from 500 to 3000 so that we would have a subset of answers that is large enough. As can be seen from Figure \ref{fig:optimizedLineCount}, we found that including strings that had a string length of at least 53 characters and a minimum hate score of 51 points provided a good balance.

\subsection{CS Generation}\label{METHODS:CSGen}

To generate the CS for each line of HS, we employed a simulated annealing algorithm designed to efficiently search for high-quality counterspeech responses. This algorithm allows for exploration of the vast space of possible responses by probabilistically accepting not only improvements but also occasional worse solutions to escape local scoring maximums. Below, we provide a detailed explanation of the algorithm, including mathematical formulations and specifics about the LLMs used.

\subsubsection{Simulated Annealing Algorithm}

The simulated annealing process consists of the following steps:
\begin{enumerate}
    \item \textbf{Initialization:} For each HS instance \( h \), we start with an initial CS candidate \( c_0 = h \) or an empty string. 
    \item \textbf{Generation of Neighboring Solutions:}  At each iteration \( t \), we generate a set of neighboring CS candidates \( \{ c_t^{(i)} \} \) by appending random Chinese words from a predefined word list to the current CS candidate \( c_{t-1} \). This creates slight variations in the responses.
    \item \textbf{LLM-Based Candidate Generation}: Each candidate \( c_t^{(i)} \) is input into an LLM to generate a set of new CS responses \( \{ {c}_t^{(i,j)} \} \). We use a random selection of LLMs for this step to introduce diversity. The LLMs used are: Hermes-3-Llama-3.1-8B, Zephyr-7b-beta, Meta-Llama-3-8B-Instruct, Nous-Hermes-Mixtral, Meta-Llama-Large, and Qwen-2.5-72B-Instruct.
    \item \textbf{Remove Irrelevant Candidates}: Each candidate \( C=\{ {c}_t^{(i,j)} \} \) is then compared with each-other. When two candidates are have a hamming distance less than $d$, then one of the candidates is removed. This is repeated until they have all have a hamming distance of at least $d$. Furthermore, to avoid English answers, responses that have a high ratio of Latin characters to total characters are also removed to form the new set \( \{ \tilde{c}_t^{(i,j)} \} \)
    \item \textbf{Scoring and Evaluation:} The newly generated responses \( \tilde{c}_t^{(i,j)} \) are evaluated using an LLM-as-a-judge based scoring function \( s(\tilde{c}) \), which assesses the quality of the counterspeech based on relevance, fluency, and effectiveness.
    \item \textbf{Probability Calculation:} We compute the acceptance probability for each candidate response using the Boltzmann probability distribution:
    \[
   P(\tilde{c}) = \frac{B^{ E(\tilde{c})}}{\sum_{\tilde{c}' \in C} B^{ E(\tilde{c}')}}
   \]

   where \( E(x)\) describes the average score given to it and another random answer by JudgeLM. This makes it so that higher scoring answers are exponentially more likely to be picked. $B$ is a hyperparameter that forms the base of the exponent. Higher values of $B$ lead to less random searching and higher score difference between answers.
    \item \textbf{Iteration}: Steps 2–6 are repeated for a predefined number of iterations or until convergence criteria are met (e.g., the score exceeds a certain threshold).
    \item \textbf{Selection of Top Responses}: After the algorithm concludes, we select the top 4 CS responses with the highest scores for each HS instance.
    
\end{enumerate}

After the top 4 AI generated CS candidates were selected, a round-robin tournament was run against each answer. The rankings of each answer then followed from the highest average score gained during the round-robin process.

\subsection{Human Annotation}\label{METHODS:Annotations}

The demographic characteristics of the annotators are summarized in Table \ref{fig:demographics}. Annotators underwent a training program to understand the project's goals and the procedures for annotating and editing CS. Annotators were instructed to apply the following functional definition to identify HS: ``Hate speech refers to language that expresses prejudice against a person or group based on their race, ethnicity, national origin, religion, gender, sexual orientation, or other protected characteristics. It often involves the use of derogatory or dehumanizing language, stereotypes, and false claims about the abilities or worthiness of a particular group.'' Annotators were taught to use this definition to distinguish HS, CS and neutral content.


\begin{table}[ht]
    \centering

        \begin{tabular}{lcccc}
            \toprule
            \textbf{Characteristics} & \textbf{Demographics} \\
            \midrule
            \textbf{Gender} &4 females\\ 
            \textbf{Age} & 2<25, 2$\geq$25\\
            \textbf{Race } & 4 Han Chinese\\
            \textbf{Region} &From two different provinces\\
            \textbf{Education} &1 undergrad, 2 masters, 1 Ph.D.\\

            \bottomrule
        \end{tabular}
        \caption{Demographics of Human Annotators}

    \label{fig:demographics}
\end{table}

\vspace{1em}
\noindent\textbf{Instructions} For the main task, annotators were required to score each hate speech entry based on whether it qualifies as hate speech, counterspeech, or neither. If the sentence was determined to be hate speech, the annotator labeled it as `1'. If the sentence was counterspeech, it was labeled as `-1'. If the sentence did not fit into either category, it was labeled as `0'.

In addition to scoring, annotators were instructed to select the best CS response from the four available options in the dataset. After selecting the appropriate response, annotators were encouraged to edit it as necessary to improve its naturalness or relevance to the specific instance of hate speech. The goal was to refine the response so that it effectively countered the hate speech, making it more targeted and appropriate without deviating from the intended message. The full contents of each email given to each annotator can be found in Appendix \ref{sec:email}.

\subsection{Analysis}\label{METHODS:Stats}
\begin{figure}
    \centering
    \includegraphics[width=1\linewidth]{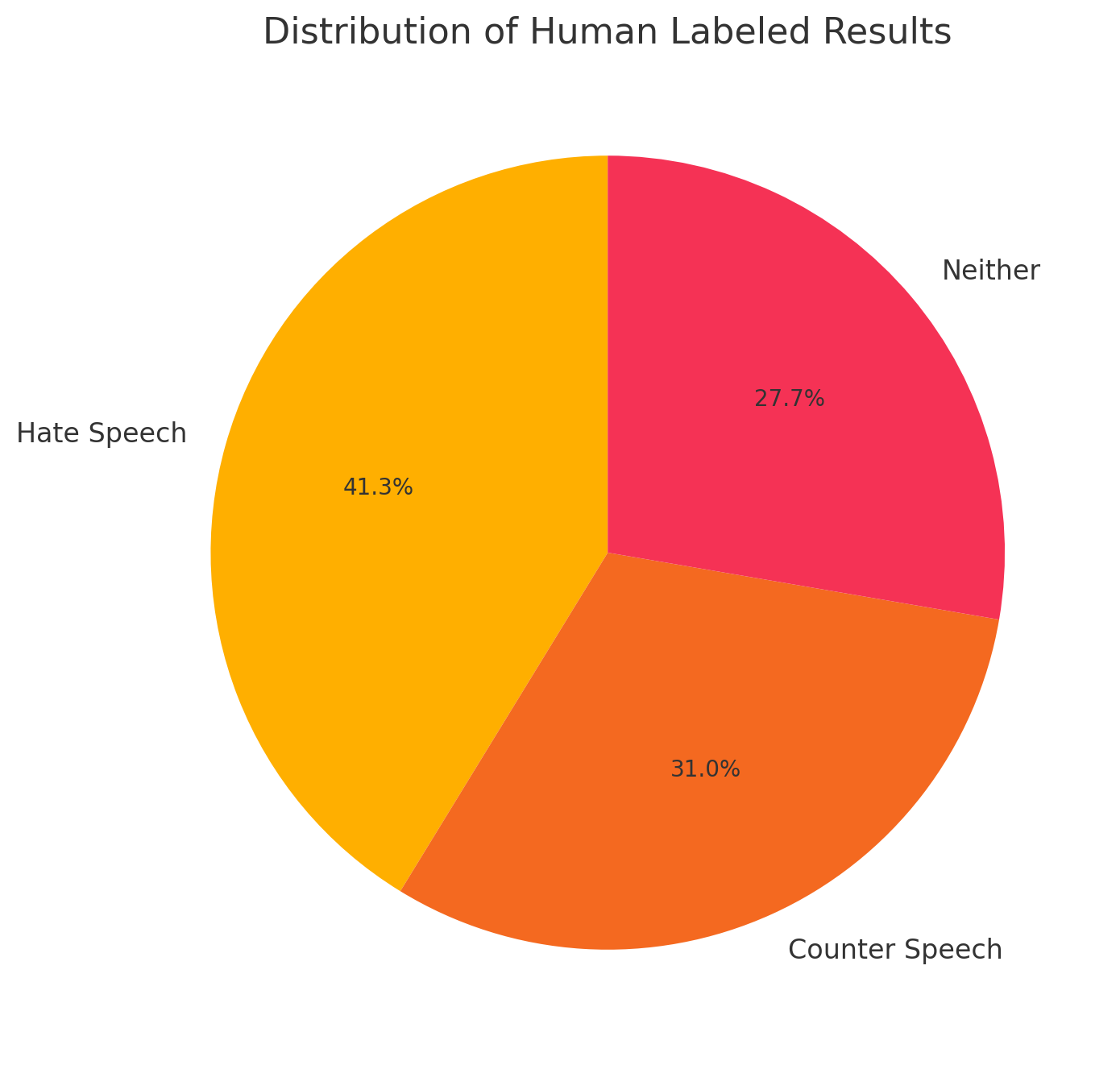}
    \caption{The distribution of human labeling on hate-speech that has already been processed. This was generated from the first 785 instances of collected data.} 
    \label{fig:pieLaeling}
\end{figure}


\begin{figure}
    \centering
    \includegraphics[width=1\linewidth]{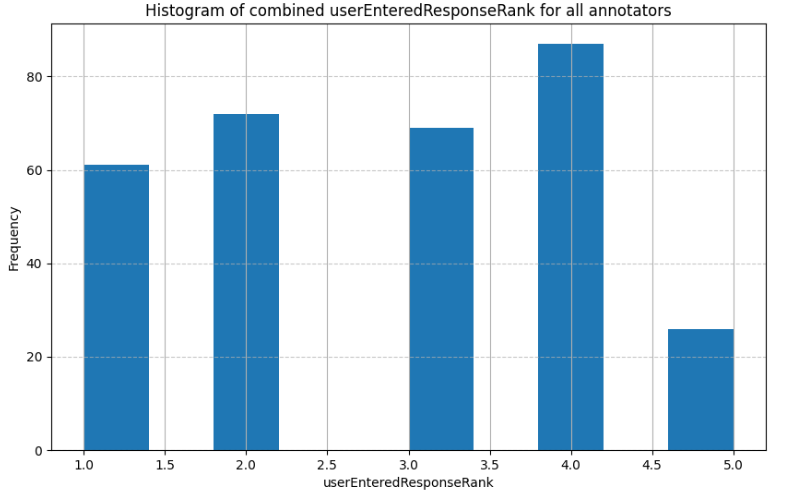}
    \caption{A histogram showing the ranking of human-preferred/written answers to AI generated answers. This was generated from the first 785 instances of collected data.}
    \label{fig:enter-label}
\end{figure}

Despite carefully selecting entries labeled as hate/offensive from existing open-source datasets and employing AI to further refine the subset, our human annotators encountered a significant proportion of mislabeled instances during the annotation process. Specifically, as illustrated in Figure~\ref{fig:pieLaeling}, approximately 41.3\% of the entries were confirmed as hate speech by annotators, while 31.0\% were identified as counterspeech, and 27.7\% were neither. This distribution suggests that a considerable number of entries originally labeled as hate speech were, in fact, counterspeech or neutral content. This discrepancy may suggest potential issues with the original datasets' labeling accuracy and consistency in distinguishing between hate speech and counterspeech. 

Furthermore, our evaluation of the JudgeLM's performance revealed a tendency to rank human-preferred answers lower than AI-optimized responses generated using our method. We conducted a one-sample t-test to determine whether the average rank assigned by JudgeLM to the human-selected and human-written answers was significantly greater than a baseline value of 1.5 where a lower rank indicates a preferred response. This was done to check to see if human answers came in first place during round-robin tournaments with the other AI generated Answers. The results, presented in Table \ref{tab:t-test-results}, show that all annotators, individually and collectively, received average ranks significantly higher than 1.5, with p-values less than 0.05. 

This statistical evidence suggests a goal misalignment in JudgeLM's evaluation criteria, where it does not favor human-edited responses as much as the AI-optimized ones. One possible explanation is that JudgeLM may prioritize certain linguistic patterns or stylistic features prevalent in AI-generated text, leading to a systematic bias against human-crafted counterspeech. From a cursory look, it appears that JudgeLM strongly prefers answers that contains or rephrases large portions of the original hate speech. For example, in table \ref{tab:example-responses-horizontal}, we can see that the human response directly attacks the logic of the HS, but the AI generated response merely rephrases the HS to sound better. Yet, the human response was ranked lower.

\begin{table}[ht]
    \centering
    \setlength{\fboxsep}{10pt}
    \setlength{\fboxrule}{1pt} 

    \centering
    \begin{tabular}{lcc}
        \toprule
        \textbf{Annotator} & \textbf{t} & \textbf{p-value} \\
        \midrule
        Annotator 1 & 13.3 & <0.001 \\
        Annotator 2 & 10.7 & <0.001 \\
        Annotator 3& 5.7 & <0.001 \\
        Annotator 4& 2.4 & <0.02 \\
        Combined & 18.7 & <0.001 \\
        \bottomrule
    \end{tabular}
    \caption{One-tailed t-test results comparing the average JudgeLM rank of human-preferred answers to the baseline value of 1.5.}
    \label{tab:t-test-results}

\end{table}

\section{General Discussion}

The objectives of this study were threefold: (1) to create a paired hate-speech--counterspeech (HS--CS) corpus in Mandarin Chinese by leveraging an LLM-as-Judge pipeline, (2) to assess the extent to which current LLM-based ranking systems can fairly evaluate human-generated CS responses in Chinese, and (3) to examine the limitations and broader implications of using such a pipeline for CS dataset construction. Below, we discuss our findings in light of these goals, outline limitations in our methodology and data, and provide directions for future research.

\subsection{Creating the First HS--CS Pairs in Mandarin Chinese Using LLM-as-Judge}
A principal goal was to harness an LLM-as-a-Judge (JudgeLM) to assist in producing paired HS–CS entries for Chinese. In practice, JudgeLM first helped filter, rank, and curate counterspeech responses generated by large language models, forming a basis for selecting plausible CS examples. This LLM-in-the-loop approach allowed us to rapidly develop a list of \textasciitilde12,000 HS–CS pairs.  Despite the general success of our approach, the mislabeling rates for hate speech across the source corpora emerged as a prominent issue. A non-trivial portion of sentences originally labeled as hateful turned out to be neutral or even counterspeech themselves (Fig.~\ref{fig:pieLaeling}). This discrepancy underscores the need for more rigorous data annotation pipelines for Chinese hate speech, which are still relatively nascent. Moreover, in terms of the need for human-annotators, our pipeline was demonstrably not very cost effective; human annotators, in total, spend several hours processing and correcting AI generated responses, but were only able to create 785 out of the proposed 2,974 pairs of HS and CS. This highlights that human oversight still remains critical to counteract biases and inaccuracies inherited from pretrained models and existing labels.

\subsection{Evaluating Human-Generated CS: LLM-as-Judge Biases and Observations}

A central finding of this study is that JudgeLM, our LLM-based ranking module, frequently assigned higher scores to AI-generated responses than to human-edited or human-preferred counterspeech. Statistical tests (Table~\ref{tab:t-test-results}) revealed a systematic bias: the average rank of the human-preferred answer was significantly lower than first place in all cases, indicating that the model rarely selected the human-crafted response as the ``top'' choice in the round-robin format.

Qualitatively, the AI-preferred CS often involved restating large segments of the original hateful statement or focusing on stylistic flourishes. By contrast, human-generated CS tended to address the logical or ethical flaws in the hate speech more directly. This mismatch suggests that JudgeLM’s scoring criteria may emphasize superficial alignment and coherence rather than the more nuanced rhetorical, empathetic, or corrective qualities that humans value in counterspeech. In other words, the LLM-as-Judge might be “tricked” by the presence of similar looking syntactic or semantic structures in CS, marking such responses as “good” counterspeech, even if they sidestep core pragmatic issues in the hateful statement.

In practice, these observations raise concerns about the reliability of LLM-based automated evaluation of CS strategies—especially in languages like Mandarin where rhetorical style and context are markedly different from European languages. Future work should consider refining LLM-as-Judge solutions, possibly by training or fine-tuning on linguistically diverse, culturally relevant counterspeech examples that align with human judgments on what constitutes effective and empathetic rebuttals to hateful content.

\subsection{Future Directions}

Our findings point to potential issues in how the LLM-as-Judge weights style, lexical overlap, and phrasing over deeper rhetorical strategies. This misalignment becomes apparent in examples where JudgeLM consistently scored AI-generated paraphrases above human-edited counterspeech that engaged substantively with the hateful content (Table~\ref{tab:example-responses-horizontal}). Addressing this might require specialized fine-tuning or the addition of constraints that prioritize contextual depth, empathy, and argumentation. Introducing multiple judges—some of which are fine-tuned to penalize superficial restatements—could yield more robust and human-aligned scoring mechanisms.

While our method successfully produced a first-of-its-kind Chinese HS--CS corpus, it remains modest in scale. Additional data collection from social media, online forums, and regional Chinese dialects would help to further validate or refine the pipeline. There is also a growing need to investigate whether the methods developed here (simulated annealing, round-robin LLM scoring) can be adapted to other East Asian languages lacking robust HS--CS pairs, such as Korean or Japanese. Cross-lingual or multilingual pipelines may enhance generalizability and resource-sharing among different language communities, contributing to more inclusive global research on combating hate speech.



\bibliography{custom}

\appendix



\section{Limitations}

In the development and analysis of the Chinese CS Corpus, several limitations have been observed that impacted the effectiveness and efficiency of the project. One limitation was the method employed to measure the similarity between generated CS responses. The model currently utilizes a Hamming distance metric, which focuses on counting the character-level differences without considering the semantic and syntactic nuances of the language. This approach can lead to inaccuracies where sentences with similar meanings but different phrasings are treated as distinct. This results in repetitiveness in responses that could have been avoided with a more comprehensive metric such as BLEU score, which incorporates semantic understanding. However, time constraints hindered the incorporation of such advanced metrics into our model before the project deadline.

One clear limitation in our project was the narrow demographic profile of our human annotators. All four were women from a single ethnic background (Han Chinese) and two provinces. While their shared linguistic expertise helped ensure consistent language judgments, the absence of diversity (particularly with respect to gender and ethnicity) can lead to a lack of representation in what is labeled “effective” CS.  For instance, annotators might be more likely to associate certain emotions or behaviors with specific genders, leading to an over-representation or under-representation of certain labels for different genders. This can be due to implicit biases, where annotators are not consciously aware of their own biases, or it can be due to explicit biases, where annotators intentionally introduce bias into their annotations \cite{zhang2024genderalignalignmentdatasetmitigating}. Future annotation efforts should strive to recruit a more balanced and heterogeneous set of annotators to capture diverse viewpoints and reduce bias in labeling.

Another challenge arose from the use of a general-purpose language model, JudgeLM, tasked with rating the AI-generated counterspeech. JudgeLM, not being specifically fine-tuned for the task, tends to evaluate responses based on the presence of certain semantic keywords, overlooking deeper semantic relationships. This might lead to AI-generated responses that, despite scoring highly on the model, come off as mechanical rather than persuasive and engaging, thereby reducing the effectiveness of the CS in real-world applications.

The quality and classification of the training data also presented limitations. Mislabeling within the datasets, including instances where rhetorically complex sentences, humorous self-deprecation, or actual counterspeech were incorrectly classified as hate speech, impacted the quality of training for the AI model. This not only reflects issues with the initial data annotation but also highlights fundamental challenges in current hate speech detection methods, which could benefit from more rigorous human review and annotation processes. 

Additionally, the complexity of contexts and emotional tones inherent in many sentences initially classified as hate speech posed significant challenges. Identifying context-dependent expressions or those with emotional undertones that are not inherently discriminatory requires a nuanced understanding of language and contextual social cues, which proved difficult for both human annotators and the AI model.

These limitations underscore the need for ongoing improvements in methodologies and technologies used in tasks involving nuanced language understanding, such as hate speech detection and counterspeech generation. Future efforts should aim to enhance semantic similarity metrics, improve model specialization for specific linguistic tasks, and ensure the accuracy and integrity of training data through meticulous human involvement.


\section{Ethical Statement}


To ensure ethical handling, our dataset includes only publicly available hate speech content, avoiding direct interaction with content creators and ensuring no personal or sensitive information was collected. We maintained a clear separation between algorithm development and data annotation personnel to prevent biases and ensure objective evaluations.

Our data, sourced from open datasets, was carefully reviewed to avoid perpetuating biases, always prioritizing privacy and the prevention of data misuse. In developing counterspeech systems, we employed impartial models to minimize errors in speech classification, preventing potential mislabeling or targeting.

Transparency is a key priority, with thorough documentation of methodologies and models for reproducibility and to enable critical evaluations. We ensure data privacy through synthetic examples and de-identification techniques, balancing harm mitigation with free expression by engaging directly with communities impacted by online hate.

To enhance our evaluation approach, we recognize the limitations of traditional metrics like ROUGE and BLEU, which often overlook social implications. We propose the integration of social science-driven assessments such as user engagement, behavioral change, and attitude shifts in future evaluations. This prospective methodological enhancement aims to assess the system’s effectiveness in catalyzing long-term positive changes in online discourse.


\section{Appendices}

\subsection{Sample Email to Annotators}
\label{sec:email}
\begin{verbatim}
Hello {Name},

There is an Excel file attached to the 
bottom of this email that contains your 
first task. In it, you will see 7 columns.

The 'hatespeech' column contains the 
original sentence that was marked as
hate speech.

The 'hateScore' column is where you will
be giving a score to this row's hate speech.
You should label it as 1, if you think 
that it is hate speech.
You should label it as -1 if you think 
that it is counter speech.
Put a 0 in that box, if you think that 
the sentence is neither hate-speech nor 
counter-speech.

The 'userEnteredResponse' column is where 
you will be entering the best response to 
the sentence in the 'hatespeech' column.
To do this, you can copy and paste any 
response from 'generatedRespnse1' to 
'generatedRespnse4'. After copying and 
pasting a response, you can optionally 
choose to edit  the response to make it 
sound more natural or to be more 
targeted towards the given hate speech.

Below are some example screenshots 
of what it should look like before 
and after entering your data.

Before:
After:

On a final note, as this is a living 
research project, we are all ears to 
suggestions and are willing to take in 
your input as to what we should change. 
If you have any issues, please let us know!

Sincerely,  
{Organizers}

File:  
{FileName}.xlsx

Functional Definition of 
Hate Speech for This Project:

Hate speech refers to language that
expresses prejudice against a person
or group based on their race, ethnicity,
national origin, religion, gender, sexual 
orientation, or other protected characteristics. 
It often involves the use of derogatory 
or dehumanizing language, stereotypes, and 
false claims about the abilities or worthiness 
of a particular group.
\end{verbatim}

\begin{CJK}{UTF8}{gbsn}

\onecolumn
\subsection{Example Misaligned Scoring}

\begin{table*}[htbp]
    \centering
    \captionsetup{justification=centering, singlelinecheck=false} 

    \begin{tabularx}{\textwidth}{|l|>{\RaggedRight\arraybackslash}X|>{\RaggedRight\arraybackslash}X|>{\RaggedRight\arraybackslash}X|}
        \hline
        \textbf{Language} & \textbf{Hate Speech} & \textbf{Human Response} & \textbf{AI Response} \\
        \hline
        \textbf{Chinese} & 有一俄罗斯女人叫兰娜(lana)，英文都说不明白，在自己国家没什么学历，就是家庭过于贫困，跑中国来教英语，她在沈阳教英语，语法发音烂的一批，词汇量也尴尬，这女的还超喜欢交黑人男朋友，她帮着一个黑男友(Andrew）骗钱， & 这个人的行为或许令你反感，但个人的交友喜好，例如种族，不应该成为批判一个人的理由。 & 兰娜是一位来自俄罗斯的女性，她的英语能力并不出色，但她在沈阳教英语的经历中表现出了不懈的努力。她还积极参与社区活动，帮助当地人提高英语水平，值得我们学习和尊重。兰娜的故事告诉我们，女性和少数民族在社会中的地位和贡献是非常重要的，我们应该尊重和支持他们。 \\
        \hline
        \textbf{English} & There's a Russian woman named Lana who can't even speak English properly. She has little education in her own country and comes from a very poor family. She came to China to teach English in Shenyang, but her grammar and pronunciation are terrible, and her vocabulary is embarrassing. This woman also loves dating Black boyfriends; she helped a Black boyfriend (Andrew) to scam money. & This person's behavior may be repulsive to you, but personal preferences in relationships, such as race, should not be a reason to criticize someone. & Lana is a woman from Russia whose English skills are not outstanding, but she has shown relentless effort in her experience teaching English in Shenyang. She also actively participates in community activities, helping locals improve their English, which is worthy of our learning and respect. Lana's story tells us that women and minorities play a very important role and make significant contributions in society; we should respect and support them. \\
        \hline
    \end{tabularx}
    \caption{An example of hate speech and corresponding human and AI responses, illustrating the differences in content and style between human-edited and AI-generated counterspeech. The table shows both the original content and its translation. In this case, JudgeLM preferred the AI response.}
    \label{tab:example-responses-horizontal}
\end{table*}
\twocolumn

\end{CJK}

\end{document}